\documentclass[10pt,twocolumn,letterpaper]{article}
\usepackage{}

\usepackage{cvpr}
\usepackage{times}
\usepackage{epsfig}
\usepackage{graphicx}
\usepackage{amsmath}
\usepackage{amssymb}
\usepackage{booktabs}

\usepackage[breaklinks=true,bookmarks=false]{hyperref}
\usepackage{algorithm}
\usepackage{algorithmic}
\usepackage{stfloats}
\usepackage{amsfonts}
\usepackage{slashbox}

\usepackage{mathrsfs}
\usepackage{booktabs}

\usepackage{multirow}

\cvprfinalcopy 

\def\cvprPaperID{****} 

\ifcvprfinal\fi

\begin{document}

\title{Fully Convolutional Adaptation Networks for Semantic Segmentation\thanks{{\small This work was performed at Microsoft Research Asia.}}}

\author{Yiheng Zhang $^{\dag}$, Zhaofan Qiu $^{\dag}$, Ting Yao $^{\ddag}$, Dong Liu $^{\dag}$, and Tao Mei $^{\ddag}$ \\
         $^{\dag}$ University of Science and Technology of China, Hefei, China\\
         $^{\ddag}$ Microsoft Research, Beijing, China\\
{\tt\small \{yihengzhang.chn, zhaofanqiu\}@gmail.com, \{tiyao, tmei\}@microsoft.com, dongeliu@ustc.edu.cn}
}

\maketitle
\thispagestyle{empty}

\begin{abstract}
The recent advances in deep neural networks have convincingly demonstrated high capability in learning vision models on large datasets. Nevertheless, collecting expert labeled datasets especially with pixel-level annotations is an extremely expensive process. An appealing alternative is to render synthetic data (e.g., computer games) and generate ground truth automatically. However, simply applying the models learnt on synthetic images may lead to high generalization error on real images due to domain shift. In this paper, we facilitate this issue from the perspectives of both visual appearance-level and representation-level domain adaptation. The former adapts source-domain images to appear as if drawn from the ``style" in the target domain and the latter attempts to learn domain-invariant representations. Specifically, we present Fully Convolutional Adaptation Networks (FCAN), a novel deep architecture for semantic segmentation which combines Appearance Adaptation Networks (AAN) and Representation Adaptation Networks (RAN). AAN learns a transformation from one domain to the other in the pixel space and RAN is optimized in an adversarial learning manner to maximally fool the domain discriminator with the learnt source and target representations. Extensive experiments are conducted on the transfer from GTA5 (game videos) to Cityscapes (urban street scenes) on semantic segmentation and our proposal achieves superior results when comparing to state-of-the-art unsupervised adaptation techniques. More remarkably, we obtain a new record: mIoU of 47.5\% on BDDS (drive-cam videos) in an unsupervised setting.
\end{abstract}

\section{Introduction}
\begin{figure}[!tb]
   \centering {\includegraphics[width=0.46\textwidth]{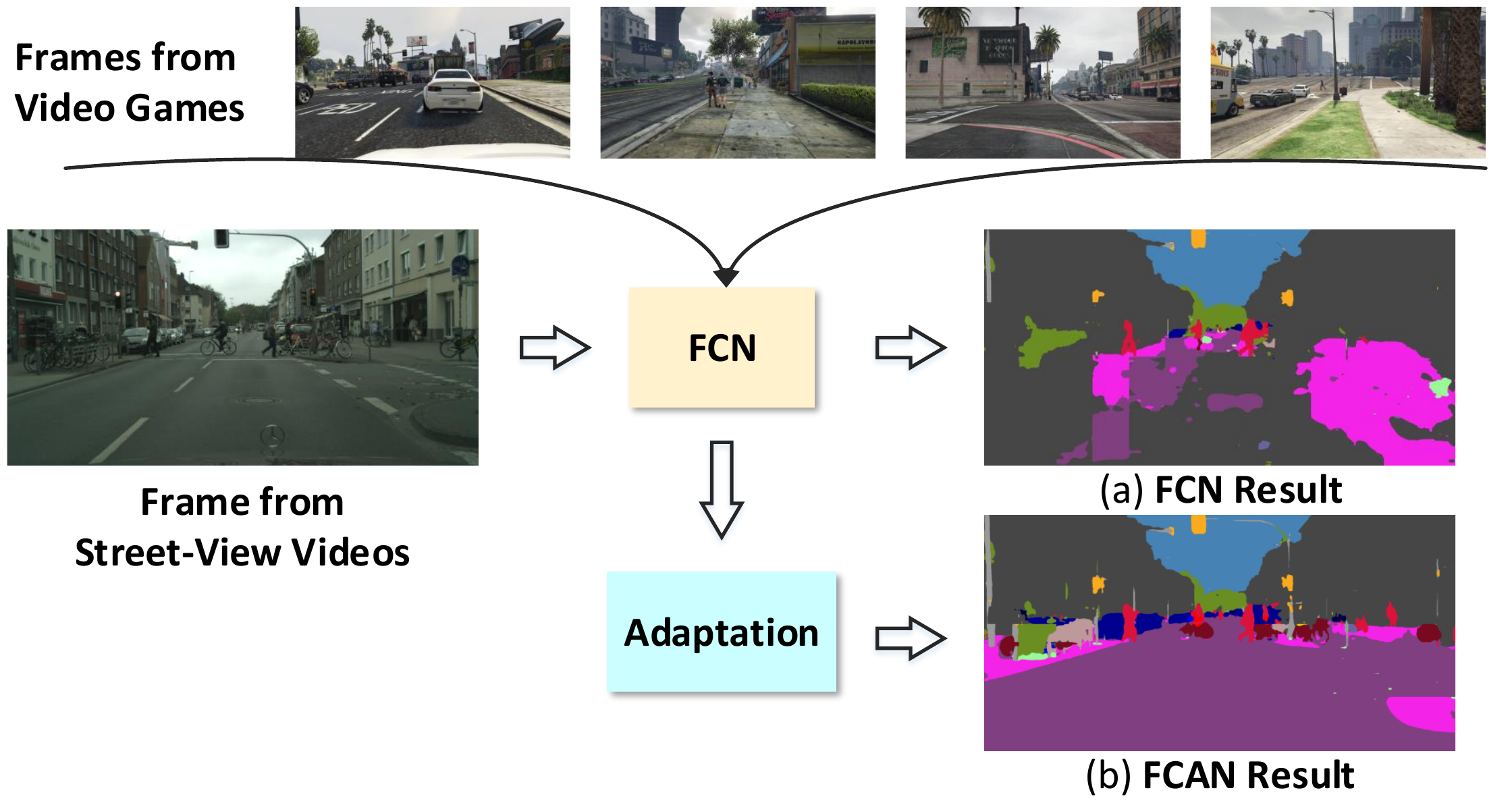}}
   \caption{\small Semantic segmentation on one example frame in street-view videos by (a) directly applying FCN trained on images from video games and (b) domain adaptation of FCAN in this work.}
   \label{intro}
   \vspace{-0.23in}
\end{figure}

Deep Neural Networks have successfully proven highly effective for learning vision models on large-scale datasets. To date in the literature, there are various datasets (e.g., ImageNet \cite{ILSVRC15} and COCO \cite{lin2014microsoft}) that include well-annotated images available for developing deep models to a variety of vision tasks, e.g., recognition \cite{he2016deep,simonyan2014very,szegedy2015going}, detection \cite{girshick2015fast,ren2015faster}, captioning \cite{Yao:ICCV17} and semantic segmentation \cite{chen2016deeplab,long2015fully}. Nevertheless, given a new dataset, the typical solution is still to perform intensive manual labeling despite expensive efforts and time-consuming process. An alternative is to utilize synthetic data which is largely available from computer games \cite{GTA5_richter2016playing} and the ground truth could be freely generated automatically. However, many previous experiences have also shown that reapplying a model learnt on synthetic data may hurt the performance in real data due to a phenomenon known as ``domain shift" \cite{Yao:CVPR15}. Take the segmentation results of one frame from real street-view videos in Figure \ref{intro} (a) as an example, the model trained on synthetic data from video games fails to properly segment the scene into semantic categories such as road, person and car. As a result, unsupervised domain adaptation would be desirable on addressing this challenge, which aims to utilize labeled examples from the source domain and a large number of unlabeled examples in the target domain to reduce a prediction error on the target~data.

A general practice in unsupervised domain adaptation is to build invariance across domains by minimizing the measure of domain shift such as correlation distances \cite{sun2016return} or maximum mean discrepancy \cite{tzeng2014deep}. We novelly consider the problem from the viewpoint of both appearance-level and representation-level invariance. The objective of appearance-level invariance is to recombine the image content in one domain with the ``style" from the other domain. As such, the images in two domains appear as if they are drawn from the same domain. In other words, the visual appearances tend to be domain-invariant. The inspiration of representation-level invariance is from the advances of adversarial learning for domain adaptation, which is to model domain distribution via an adversarial objective with respect to a domain discriminator. The spirit behind is from generative adversarial learning \cite{Goodfellow:NIPS14}, that trains two models, i.e., a generative model and a discriminative model, by pitting them against each other. In the context of domain adaptation, this adversarial principle is then equivalent to guiding the representation learning in both domains, making the difference between source and target representation distributions indistinguishable through the domain discriminator. We follow this elegant recipe and capitalize on adversarial mechanism to learn image representation that is invariant across domains. In this work, we are particularly investigating the problem of domain adaptation on semantic segmentation task which relies on probably the most accurate pixel-level annotations.

\begin{figure*}
    \begin{center}
        \includegraphics[width=0.95\linewidth]{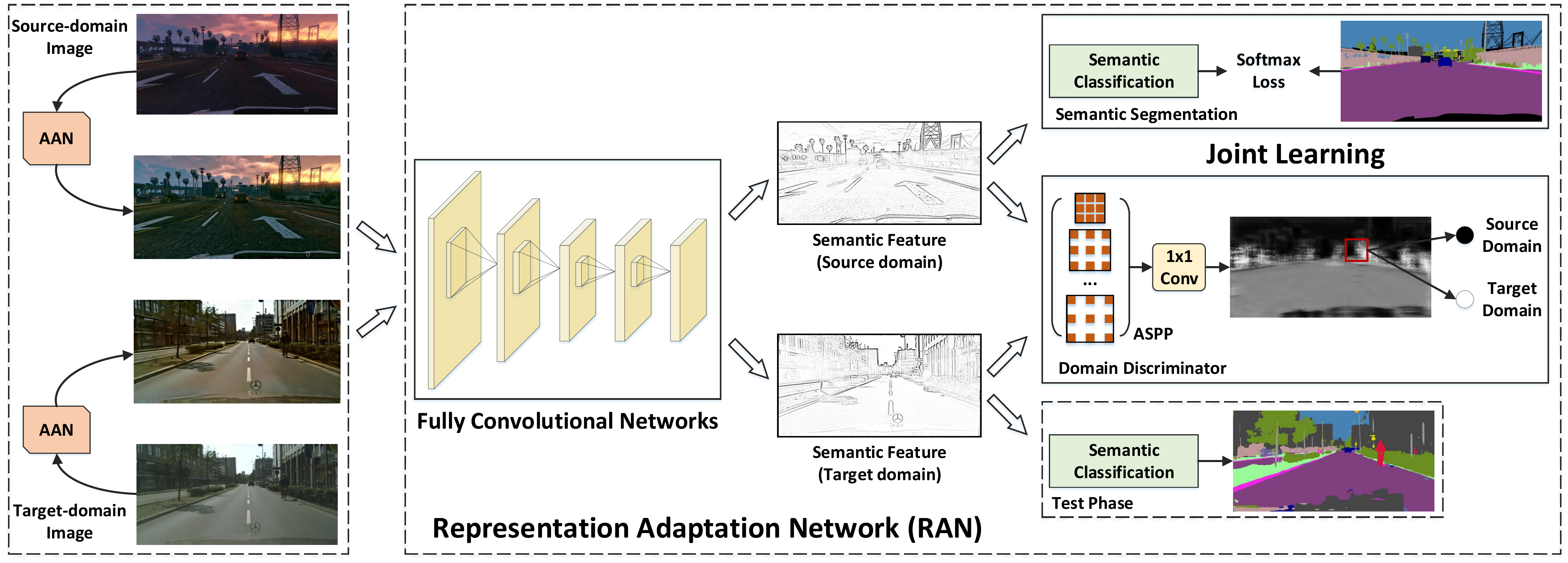}
    \end{center}
    \vspace{-0.13in}
    \caption{\small An overview of our Fully Convolutional Adaptation Networks (FCAN) architecture. It consists of two main components: the Appearance Adaptation Networks (AAN) on the left and the Representation Adaptation Networks (RAN) on the right. AAN transfers images from one domain to the other one and thus the visual appearance tends to be domain-invariant. RAN learns domain-invariant representations in an adversarial manner by maximally fooling the domain discriminator with the learnt source and target representations. An extended Atrous Spatial Pyramid Pooling (ASPP) layer is particularly devised to leverage the regions across different scales for enhancing the discriminative capability. RAN is jointly optimized with supervised segmentation loss on source images plus adversarial loss.}
    \label{fig:Framework}
    \vspace{-0.22in}
\end{figure*}

By consolidating the idea of appearance-level and representation-level invariance into unsupervised domain adaption for enhancing semantic segmentation, we present a novel Fully Convolutional Adaptation Networks (FCAN) architecture, as shown in Figure \ref{fig:Framework}. The whole framework consists of Appearance Adaptation Networks (AAN) and Representation Adaptation Networks (RAN). Ideally, AAN is to construct an image that captures high-level content in a source image and low-level pixel information of the target domain. Specifically, AAN starts with a white noise image and adjusts the output image by using gradient descent to minimize the Euclidean distance between the feature maps of the output image and those of the source image or mean feature maps of the images in target domain. In RAN, a shared Fully Convolutional Networks (FCN) is first employed to produce image representation in each domain, followed by bilinear interpolation to upsample the outputs for pixel-level classification, and meanwhile a domain discriminator to distinguish between source and target domain. An Atrous Spatial Pyramid Pooling (ASPP) strategy is particularly devised to enlarge the field of view of filters in feature map and endow the domain discriminator with more power. RAN is trained by optimizing two losses, i.e., classification loss to measure pixel-level semantics and adversarial loss to maximally fool the domain discriminator with the learnt source and target representations. With both appearance-level and representation-level adaptations, our FCAN could better build invariance across domains and thus obtain encouraging segmentation results in Figure \ref{intro} (b).

The main contribution of this work is the proposal of Fully Convolutional Adaptation Networks for addressing the issue of semantic segmentation in the context of domain adaptation. The solution also leads to the elegant views of what kind of invariance should be built across domains for adaptation and how to model the domain invariance in a deep learning framework especially for the task of semantic segmentation, which are problems not yet fully understood in the literature.

\section{Related Work}
We briefly group the related works into two categories: semantic segmentation and deep domain adaptation.

\textbf{Semantic segmentation} is one of the most challenging tasks in computer vision, which attempts to predict pixel-level semantic labels of the given image or video frame. Inspired by the recent advance of Fully Convolutional Networks (FCN) \cite{long2015fully}, there have been several techniques, ranging from multi-scale feature ensemble (e.g., Dilated Convolution \cite{DilatedConv_yu2015multi}, RefineNet \cite{Lin:2017:RefineNet}, DeepLab \cite{chen2016deeplab} and HAZNet \cite{xia2016zoom}) to context information preservation (e.g., ParseNet \cite{liu2015parsenet}, PSPNet \cite{zhao2017pspnet} and DST-FCN \cite{Qiu:TMM17}), being proposed. The original FCN formulation could also be improved by exploiting some post processing techniques (e.g., conditional random fields \cite{zheng2015conditional}). Moreover, as most semantic segmentation methods rely on the pixel-level annotations which require extremely expensive labeling efforts, researchers have also strived to leverage weak supervision instead (e.g., instance-level bounding boxes~\cite{dai2015boxsup}, image-level tags~\cite{pinheiro2015image}) for semantic segmentation task. To achieve this target, the techniques such as multiple instance learning~\cite{pathak2014fully}, EM algorithm~\cite{papandreou2015weakly} and constrained CNN~\cite{pathak2015constrained} are exploited in the literature. An alternative in \cite{hong2016learning} utilizes the pixel-level annotations from auxiliary categories to generalize semantic segmentation to categories where only image-level labels are available. The goal of this work is to study the exploration of freely accessible synthetic data with annotations and largely unlabeled real data for annotating real images on the pixel level, which is an emerging research area.

\textbf{Deep Domain adaptation} aims to transfer model learnt in a labeled source domain to a target domain in a deep learning framework. The research of this topic has proceeded along three different dimensions: unsupervised adaptation, supervised adaptation and semi-supervised adaptation.
Unsupervised domain adaptation refers to the setting when the labeled target data is not available. Deep Correlation Alignment (CORAL)~\cite{sun2016return} exploits Maximum Mean Discrepancy (MMD) to match the mean and covariance of source and target distributions.
Adversarial Discriminative Domain Adaptation (ADDA) \cite{tzeng2017adversarial} optimizes the adaptation model with adversarial training.
In contrast, when the labeled target data is available, we refer to the problem as supervised domain adaptation.
Tzeng \emph{et al.}~\cite{tzeng2015simultaneous} utilizes a binary domain classifier and devises the domain confusion loss to encourage the predicted domain labels to be uniformly distributed.
Deep Domain Confusion (DDC)~\cite{tzeng2014deep} applies MMD as well as the regular classification loss on the source to learn representations that are both discriminative and domain invariant.
In addition, semi-supervised domain adaptation methods have also been proposed, which exploit both labeled and unlabeled target data.
Deep Adaptation Network (DAN)~\cite{long2015learning} embeds all task specific layers in a reproducing kernel Hilbert space. Both semi-supervised and unsupervised settings are considered.

In short, our work in this paper mainly focuses on unsupervised adaptation for semantic segmentation task, which is seldom investigated. The most closely related work is the FCNWild~\cite{BDDS_hoffman2016fcns}, which addresses the cross-domain segmentation problem by only exploiting fully convolutional adversarial training for domain adaptation.
Our method is different from~\cite{BDDS_hoffman2016fcns} in that we solve the domain shift from the perspectives of both visual appearance-level and representation-level domain adaptation, which bridges the domain gap in a more principled way.

\section{Fully Convolutional Adaptation Networks (FCAN) for Semantic Segmentation}
\label{FCAN}
In this section we present our proposed Fully Convolutional Adaptation Networks (FCAN) for semantic segmentation.
Figure \ref{fig:Framework} illustrates the overview of our framework.
It consists of two main components: the Appearance Adaptation Networks (AAN) and the Representation Adaptation Networks (RAN).
Given the input images from two domains, AAN is first utilized to transfer images from one domain to the other from the perspective of visual appearance.
By recombining the image content in one domain with the ``style'' from the other one, the visual appearance tends to be domain-invariant.
We take the transformation from source to target as an example in this section, and the other options will be elaborated in our experiments.
On the other hand, RAN learns domain-invariant representations in an adversarial manner and a domain discriminator is devised to classify which domain the image region corresponding to the receptive field of each spatial unit in the feature map comes from. The objective of RAN is to guide the representation learning in both domains, making the source and target representations indistinguishable through the domain discriminator. As a result, our FCAN addresses domain adaptation problem from the viewpoint of both visual appearance-level and representation-level domain invariance and is potentially more effective at undoing the effects of domain shift.

\subsection{Appearance Adaptation Networks (AAN)}
The goal of AAN is to make the images from different domains visually similar. In other words, AAN tries to adapt the source images to appear as if drawn from the target domain. To achieve this, the low-level features over all the images in target domain should be separated and regarded as the ``style" of target domain, as these features encode the low-level forms of the images, e.g., texture, lighting and shading. In contrast, the high-level content in terms of objects and their relations in the source image should be extracted and recombined with the ``style" of target domain to produce an adaptive image.

\begin{figure}
\begin{center}
\includegraphics[width=0.98\linewidth]{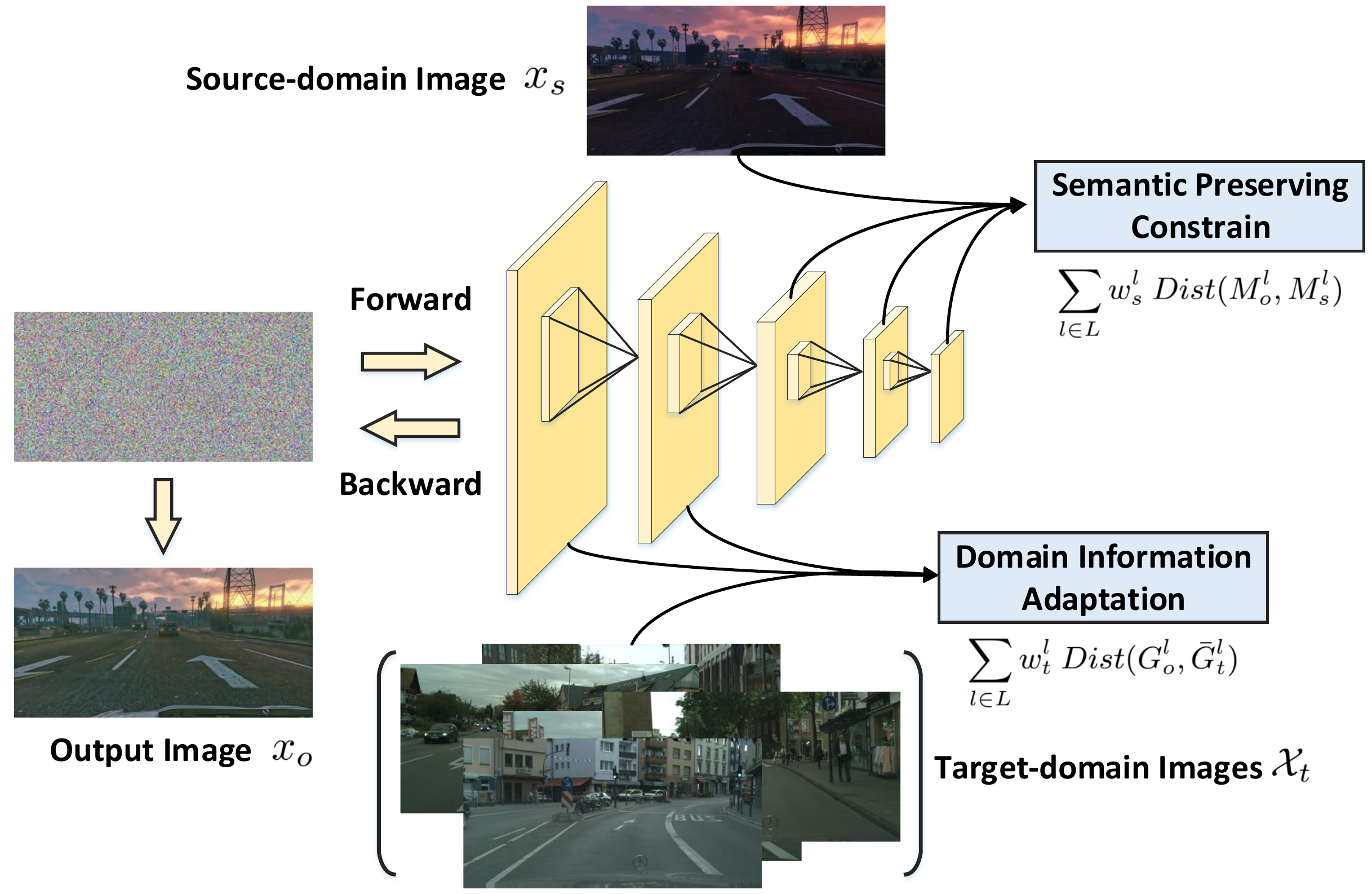}
\end{center}
\vspace{-0.06in}
\caption{\small The architecture of Appearance Adaptation Networks (AAN). Given the target image set $\mathcal{X}_t$ and one source image $x_s$, we begin with a white noise image and adjust it towards an adaptive image $x_o$, which appears as if it is drawn from target domain but contains semantic content in the source image. A pre-trained CNN is utilized to extract feature maps. The high-level image content of $x_s$ is preserved by minimizing the distance between feature maps of $x_s$ and $x_o$, while the style of target domain is kept by minimizing the distance between feature correlations of $x_o$ and~$\mathcal{X}_t$.}
\label{fig:AAN}
 \vspace{-0.2in}
\end{figure}

Figure \ref{fig:AAN} illustrates the architecture of AAN.
Given a set of images $\mathcal{X}_{t}=\{x_t^{i}|i=1,\dots,m\}$ in target domain and one image from source domain $x_{s}$, we begin with a white noise image and iteratively render this image with the semantic content in $x_{s}$ plus the ``style" of $\mathcal{X}_{t}$ to produce an adaptive image $x_o$.
Specifically, a pre-trained CNN is utilized to extract feature maps for each image. Suppose every convolutional layer $l$ in the CNN has $N_l$ response maps, where $N_l$ is the number of channels, and the size of each response map is $H_l\times W_l$, where $H_l$ and $W_l$ denotes the height and width of the map, respectively. As such, the feature maps in the layer $l$ could be represented as $M^l\in\mathbb{R}^{N_l\times H_l\times W_l}$. Basically the responses in different convolutional layers characterize image content on different semantic level, where deeper layer responds to higher semantics. To better govern the semantic content in source image $x_s$, different weights are assigned to different layers to reflect the contribution of each layer. The objective function is then formulated as
\begin{equation}
\small
\label{loss_appearance_1}
\begin{aligned}
\min_{x_o} \sum_{l\in L} w_{s}^{l}~Dist(M_o^l, M_s^l)~,
\end{aligned}
\end{equation}
where $L$ is the set of layers to be considered for measurement. $w_{s}^{l}$ is the weight of layer $l$, $M_o^l$ and $M_s^l$ is the feature map of layer $l$ on $x_o$ and $x_s$, respectively. By minimizing the Euclidean distance in Eq.(\ref{loss_appearance_1}), the image content in $x_s$ is expected to be preserved in the adaptive image $x_o$.

Next, the ``style" of one image is in general treated as a kind of statistical measurement or \emph{pattern}, which is agnostic to spatial information \cite{gatys2015texture}. In CNN, one of such statistical measurements is the correlations between different response maps. Hence, the ``style" of an image $G^l$ on layer $l$ could be computed by
\begin{equation}
\small
\label{loss_appearance_2}
\begin{aligned}
{G^{l,ij}}={M^{l,i}\odot M^{l,j}}~.
\end{aligned}
\end{equation}
$G^{l,ij}$ is the inner product between the vectorized $i$-th and $j$-th response map of $M^l$.
In our case, we extend the ``style" of one image to that of one domain ($\bar G_{t}^l$ of the target domain) by averaging $G^l$ over all the images in target domain. In order to synthesize the ``style" of target domain into $x_o$, we formulate the objective in each layer as
\begin{equation}
\small
\label{loss_appearance_3}
\begin{aligned}
\min_{x_o} \sum_{l\in L} w_{t}^{l}~Dist(G_o^l, \bar G_t^l)~,
\end{aligned}
\end{equation}
where $w_{t}^{l}$ is the weight for layer $l$. Finally, the overall loss function $\mathcal{L}_{AAN}$ to be minimized is
\begin{equation}
\small
\label{loss_appearance_4}
\begin{aligned}
\mathcal{L}_{AAN}(x_o)=\sum_{l\in L} w_{s}^{l}~ Dist(M_o^l, M_s^l) + \alpha\sum_{l\in L} w_{t}^{l}~Dist(G_o^l, \bar G_t^l)~,
\end{aligned}
\end{equation}
where $\alpha$ is the weight to balance semantic content in the source image and the style of target domain.
In the training, similar to \cite{gatys2016image}, AAN adjusts the output image by back-propagating the gradients derived from Eq.~(\ref{loss_appearance_4}) to $x_o$, resulting in the domain-invariant appearance.

\subsection{Representation Adaptation Networks (RAN)}
\label{RAN}
With the Appearance Adaptation Networks, the images from different domains appear to be from the same domain. To further reduce the impact of domain shift, we attempt to learn domain-invariant representations. Consequently, Representation Adaptation Networks (RAN) is designed to adapt representations across domains, which is derived from the idea of adversarial learning \cite{Goodfellow:NIPS14}. The adversarial principle in our RAN is equivalent to guiding the learning of feature representations in both domains by fooling a domain discriminator $D$ with the learnt source and target representations. Specifically, RAN first utilizes a shared Fully Convolutional Network (FCN) to extract the representations of images or adaptive images through AAN from both domains. This FCN model $F$ here aims to learn indistinguishable image representations across two domains. Furthermore, the discriminator $D$ attempts to differentiate between source and target representations, whose outputs are the domain prediction of each image region that corresponds to the spatial unit in the final feature map. Formally, given the training set $\mathcal{X}_{s}=\{x_s^{i}|i=1,\dots,n\}$ in source domain and $\mathcal{X}_{t}=\{x_t^{i}|i=1,\dots,m\}$ in target domain, the adversarial loss $\mathcal{L}_{adv}$ is the average classification loss over all spatial units, which is formulated as
\begin{equation} \small
\label{loss_adv}
\begin{split}
&\mathcal{L}_{adv}(\mathcal{X}_{s},\mathcal{X}_{t})=-E_{x_t \sim \mathcal{X}_t}[\frac{1}{Z}\sum_{i=1}^{Z}log(D_{i}(F(x_t)))] \\
&\quad \quad \quad \quad \quad\quad~ -E_{x_s \sim \mathcal{X}_s}[\frac{1}{Z}\sum_{i=1}^{Z}log(1-D_{i}(F(x_s)))]~,
\end{split}
\end{equation}
where $Z$ is the number of spatial units in the output of $D$.
Similar to the standard GANs, the adversarial training of our RAN is to optimize the following minimax function
\begin{equation} \small \label{loss_adv_2}
\begin{split}
\max_{F}\min_{D}{\mathcal{L}_{adv}(\mathcal{X}_{s},\mathcal{X}_{t})}~.
\end{split}
\end{equation}

Given the fact that there are many different objects of various size in real data, we further take the utilization of multi-scale representations into account to enhance the adversarial learning.
One traditional multi-scale strategy is to resize the images with multiple resolutions, which indeed improves the performance but at the cost of large computation.
In this work, we extend Atrous Spatial Pyramid Pooling (ASPP) \cite{chen2016deeplab} to implement this, as shown in Figure \ref{fig:Framework}.
Specifically, $k$ dilated convolutional layers with different sampling rates are exploited in parallel to produce $k$ feature representations on the output of FCN independently, each with $c$ feature channels. All the feature channels are then stacked up to form a new feature map with $ck$ channels, followed by a $1\times1$ convolutional layer plus a sigmoid layer to generate the final score map. Each spatial unit in the score map presents the probability of the corresponding image region belonging to the target domain. In addition, we simultaneously optimize the standard pixel-level classification loss $\mathcal{L}_{seg}$ for supervised segmentation on the images from source domain, where the labels are available.
Hence, the overall objective of RAN integrates $\mathcal{L}_{seg}$ and $\mathcal{L}_{adv}$ as
\begin{equation} \small
\label{loss_adv_3}
	\max_{F}\min_{D}
	\left\{
	\mathcal{L}_{adv}(\mathcal{X}_{s},\mathcal{X}_{t}) - \lambda\mathcal{L}_{seg}(\mathcal{X}_{s})
	\right\}~,
\end{equation}
where $\lambda$ is the tradeoff parameter. Through fooling the domain discriminator with the source and target representations, our RAN is able to produce domain-invariant representations. In test stage, the images in target domain are fed into the learnt FCN to produce representations for pixel-level classification.

\section{Implementation}
\subsection{Appearance Adaptation}
We adopt the pre-trained ResNet-50~\cite{he2016deep} architecture as the basic CNN. In particular, we only include the five convolutional layers in the set, i.e., $L=\{conv1, res2c, res3d, res4f, res5c\}$, as the representations of these layers in general have the highest capability in each scale. The weights $w_s^{l}$ and $w_t^{l}$ of layers for the images in source and target domain are generally determined on the visual appearances of adaptive images. In addition, when optimizing Eq.~(\ref{loss_appearance_4}), a common problem is the need to set the tradeoff parameter $\alpha$ to balance content and ``style." As the ultimate goal is to semantically segment each pixel in the images, it is required to preserve the semantic content precisely. As a result, the impact of ``style" is regarded as only a ``delta" function to adjust the appearance and we empirically set a small weight of $\alpha=10^{-14}$ for this purpose. The number of maximum iteration $I$ is fixed to 1$k$. In each iteration $i$, the image $x_o$ is updated by $x_o^{i}=x_o^{i-1}-w^{i-1}\frac{g^{i-1}}{\|g^{i-1}\|_1}$, where $\tiny{g^{i-1}=\frac{\partial \mathcal{L}_{app}(x_o^{i-1})}{\partial x_o^{i-1}}}$, ${w^{i-1}=\beta\frac{I-i}{I}}$ and $\beta=10$.

\subsection{Representation Adaptation}
In our implementations, we employ dilated fully convolutional network~\cite{chen2016deeplab} originated from ResNet-101~\cite{he2016deep} as our FCN, which has proven to be effective on generating powerful representations for semantic segmentation.
The feature maps of the last convolutional layer (i.e., $res5c$) are fed into both segmentation and adversarial branches.
In supervised segmentation branch, we also augment the outputs of FCN with Pyramid Pooling \cite{zhao2017pspnet} to integrate contextual prior into representation.
In adversarial branch, we use $k=4$ dilated convolutional layers in parallel to produce multiple feature maps, each with $c=128$ channels. The sampling rate of different dilated convolution kernel is $1,2,3$ and $4$, respectively.
Finally, a sigmoid layer is utilized next to the ASPP to output the predictions, which are in the range of $[0, 1]$.

\subsection{Training Strategy}
Our proposal is implemented on Caffe~\cite{jia2014caffe} framework and mini-batch stochastic gradient descent algorithm is exploited to optimize the model. We pre-train RAN on source domain with only segmentation loss. The initial learning rate is 0.0025. Similar to~\cite{chen2016deeplab}, we use the ``poly'' learning rate policy with power fixed to 0.9. Momentum and weight decay is set to 0.9 and 0.0005, respectively. The batch size is 6. The maximum iteration number is 30$k$. Then, we fine-tune RAN jointly with segmentation loss and adversarial loss. The tradeoff parameter $\lambda$ is set to 5. The initial learning rate is 0.0001. The batch size is 8 and the maximum iteration number is 10$k$. The rest hyper-parameters are the same with those in pre-training.

\section{Experiments}
\subsection{Datasets}
We conduct a thorough evaluation of our FCAN on the domain adaptation from GTA5 \cite{GTA5_richter2016playing} (game videos) dataset to Cityscapes (urban street scenes) dataset \cite{Cordts2016Cityscapes}.

The GTA5 dataset contains 24,966 images (video frames) from the game Grand Theft Auto V (GTA5) and the pixel-level ground truth for each image is also created. In the game, the images are captured on the virtual city of Los Santos, which is originated from the city of Los Angeles. The resolution of each image is $1914 \times 1052$. There are 19 classes which are compatible with other segmentation datasets for outdoor scenes (e.g., Cityscapes) and utilized in the evaluation. The Cityscapes dataset is one popular benchmark for semantic understanding of urban street scenes, which contains high quality pixel-level annotations of 5,000 images (frames) collected in street scenes from 50 different cities. The image resolution is $2048 \times 1024$. Following the standard protocol in segmentation task (e.g., \cite{Cordts2016Cityscapes}), 19 semantic labels (car, road, person, building, etc.) are used for evaluation. In between, the training, validation, and test sets contains 2,975, 500, and 1,525 frames, respectively. Following the settings in \cite{BDDS_hoffman2016fcns,peng2017visda}, only the validation set (500 frames) are exploited for validating the unsupervised semantic segmentation in our experiments.

In addition, we also take the Berkeley Deep Driving Segmentation (BDDS) dataset~\cite{BDDS_hoffman2016fcns} as another target domain for verifying the merit of our FCAN. The BDDS dataset consists of thousands of dashcam video frames with pixel-level annotations, which share compatible label space with Cityscapes. The image resolution is $1280 \times 720$. Following the settings in \cite{BDDS_hoffman2016fcns,peng2017visda}, 1,500 frames are used for evaluation.

In all experiments, we adopt the Intersection over Union (IoU) per category and mean IoU over all the categories as the performance metrics.

\begin{table}
    \centering
    \small
    \caption{\small The mIoU performance comparisons between different ways of utilizing AAN.}
    \begin{tabular}{l|l|c|c} \hline
    \textbf{Train} & \textbf{Validation} & \textbf{FCN} & \textbf{RAN}  \\ \hline
    Src          & Tar         & 29.15  & 44.81        \\\hline
    Src          & Tar\_Ada    & 34.68  & 45.03        \\\hline
    Src\_Ada     & Tar         & 31.71  & \textbf{46.21}     \\\hline
    Src\_Ada     & Tar\_Ada    & 36.25  & 45.59        \\\hline \hline
    \multicolumn{2}{c|}{Late Fusion} & 37.61 & \textbf{46.60}          \\\hline
    \end{tabular}
    \label{tab:AAN}
    \vspace{-0.07in}
\end{table}

\begin{figure}[t]
    \begin{center}
        \includegraphics[width=1.0\linewidth]{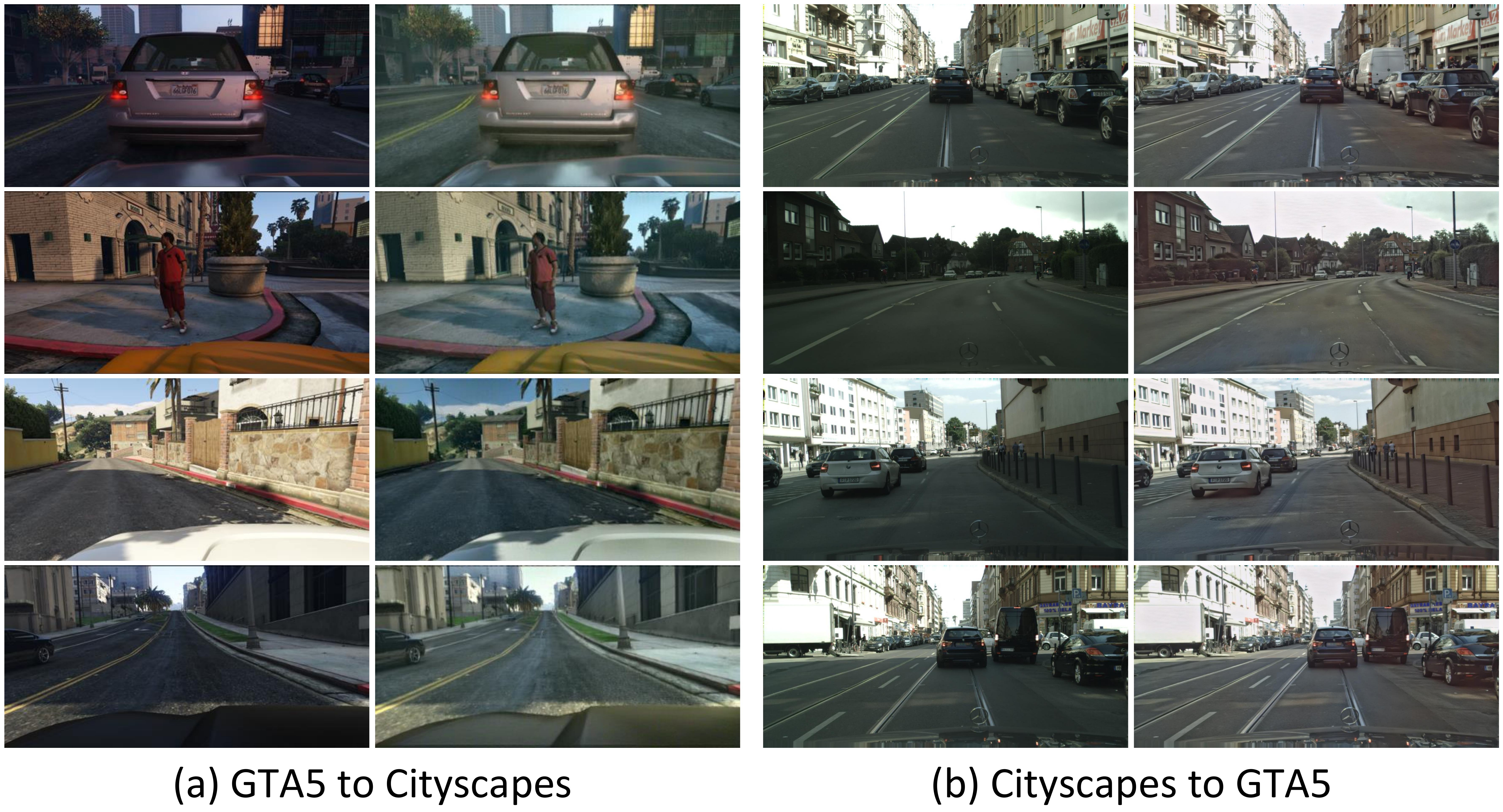}
    \end{center}
    \vspace{-0.1in}
    \caption{\small Examples of appearance-level adaptation through AAN.}
    \label{fig:Image Adapation Visual}
    \vspace{-0.2in}
\end{figure}

\subsection{Evaluation of AAN}
We first examine the effectiveness of AAN on semantic segmentation from two aspects: 1) images from which domain are adapted by AAN, and 2) adaptation by only performing AAN or plus RAN. Source Adaptation (Src\_Ada) here is to render source images with the ``style" of the target domain, and vice versa for Target Adaptation (Tar\_Ada). \textbf{FCN} refers to the setting of semantic segmentation by directly exploiting the FCN learnt on source domain to do prediction on target images. In contrast, \textbf{RAN} further performs representation-level adaptation by our RAN.

The mIoU performances between different ways of utilizing AAN are summarized in Table \ref{tab:AAN}. Overall, adapting images in source domain through AAN plus RAN achieves the highest mIoU of 46.21\%. The results by applying AAN to images in source or target or both domains consistently exhibits better performance than the setting without the use of AAN (the first row) when directly employing FCN in segmentation. The results basically indicate the advantage of exploring appearance-level domain adaptation. The performance in each setting is further improved by RAN, indicating that visual appearance-level and representation-level adaptation are complementary to each other. Another observation is that the performance gain of RAN tends to be large when performing AAN on source images. The gain is however decreased when adapting target images by AAN. We speculate that this may be the result of synthesizing some noise into the adapted target images by AAN especially at the boundary of objects and that in turn affects the segmentation stability. Furthermore, when late fusing the score maps of segmentation predicted by the four settings, the mIoU performance could be boosted up to 46.6\%. We refer to this fusion version as AAN in the following evaluations unless otherwise stated.

Figure \ref{fig:Image Adapation Visual} shows four examples of appearance-level transfer for images in source and target domain, respectively. As illustrated in the figure, the semantic content in original images are all well-preserved in the adaptive images. When rendering the images in GTA5 with ``style" of Cityscapes, the overall color of the images becomes bleak and the color saturation tends to be low. In contrast, when reversing the transfer direction, the color of images in Cityscapes gets much brighter and with high saturation. The results demonstrate a good appearance-level transfer in between.

\begin{table}
    \centering
    \small
    \caption{\small Performance contribution of each design in FCAN.}
    \begin{tabular}{l|ccccc|c} \hline
        \textbf{Method} & \textbf{ABN} & \textbf{ADA} & \textbf{Conv} & \textbf{ASPP} & \textbf{AAN} & \textbf{mIoU} \\ \hline
        FCN & & & & & & 29.15 \\\hline
        +ABN & $\surd$ & & & & & 35.51 \\\hline
        +ADA & $\surd$ & $\surd$ & & & & 41.29 \\\hline
        +Conv& $\surd$ & $\surd$ & $\surd$ & & & 43.17 \\\hline
        +ASPP& $\surd$ & $\surd$ & $\surd$ & $\surd$ & & 44.81 \\\hline \hline %
        FCAN & $\surd$ & $\surd$ & $\surd$ & $\surd$ & $\surd$ & \textbf{46.60} \\\hline
    \end{tabular}
    \label{tab:Methods Effects}
    \vspace{-0.25in}
\end{table}

\begin{figure*}[t]
    \begin{center}
        \includegraphics[width=1.0\linewidth]{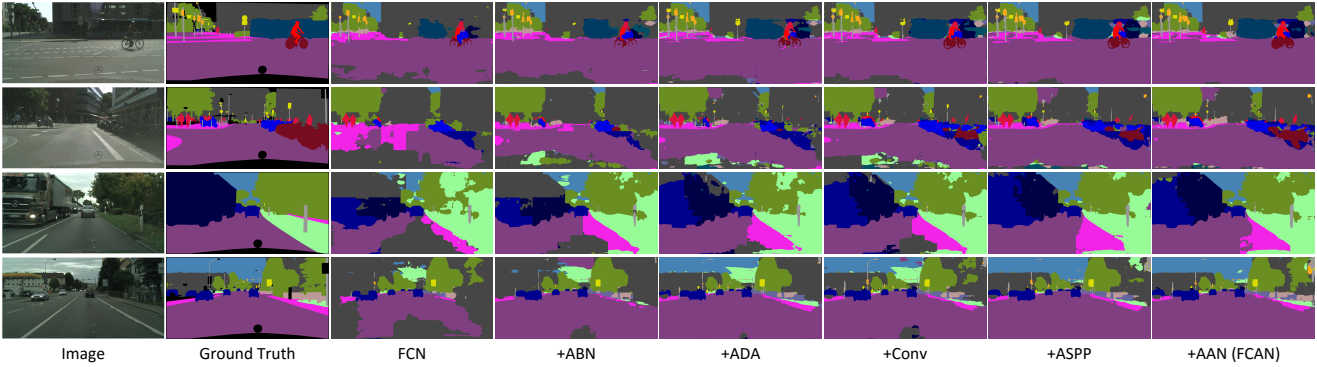}
    \end{center}
     \vspace{-0.17in}
    \caption{\small Examples of semantic segmentation results in Cityscapes. The original images, their ground truth and comparative segmentation results at different stages of FCAN are given.}
    \label{fig:Methods Effect Visual}
       \vspace{-0.2in}
\end{figure*}

\subsection{An Ablation Study of FCAN}
Next, we study how each design in FCAN influences the overall performance. Adaptive Batch Normalization (\textbf{ABN}) simply replaces the mean and variance of BN layer in FCN learnt in source domain with those computed on the images in target domain. Adversarial Domain Adaptation (\textbf{ADA}) leverages the idea of adversarial training to learn domain-invariant representations and the domain discriminator judges the domain on image level. When the domain discriminator is extended to classify each image region, this design is named as \textbf{Conv}. \textbf{ASPP} further enlarges the field of view of filters to enhance the adversarial learning. \textbf{AAN} is our appearance-level adaptation.

Table \ref{tab:Methods Effects} details the mIoU improvement by considering one more factor for domain adaptation at each stage in FCAN. ABN is a general way to alleviate domain shift irrespective of any domain adaptation frameworks. In our case, ABN successfully brings up the mIoU performance from 29.15\% to 35.51\%. This demonstrates that ABN is a very effective and practical choice. ADA, Conv and ASPP are three specific designs in our RAN and the performance gain of each is 5.78\%, 1.88\% and 1.64\%, respectively. In other words, our RAN leads to a large performance boost of 9.3\% in total. The results verify the idea of representation-level adaptation. AAN further contributes an mIoU increase of 1.79\% and the mIoU performance of FCAN finally reaches 46.6\%. Figure \ref{fig:Methods Effect Visual} showcases four examples of semantic segmentation results at different stages of our FCAN. As illustrated in the figure, the segmentation results are becoming increasingly accurate as more adaptation designs are included. For instance, at the early stages, the majority categories such as road and sky cannot be well segmented. Instead, even the minority classes such as bicycle and truck are segmented nicely during the latter steps.

\begin{table}
    \centering
    \small
    \caption{\small Performance comparisons with the state-of-the-art unsupervised domain adaptation methods on Cityscapes.}
    \begin{tabular}{l|c} \hline
        \textbf{Method} & \textbf{mIoU} \\ \hline
        DC~\cite{tzeng2015simultaneous} & 37.64 \\
        ADDA~\cite{tzeng2017adversarial}             & 38.30 \\
        FCNWild~\cite{BDDS_hoffman2016fcns}          & 42.04 \\ \hline \hline
        FCAN                                         & \textbf{46.60} \\
        FCAN(MS)                                     & \textbf{47.75}   \\ \hline
    \end{tabular}
    \label{tab:state-of-the-art}
    \vspace{-0.23in}
\end{table}

\subsection{Comparisons with State-of-the-Art}
\begin{figure}[t]
    \begin{center}
        \includegraphics[width=1.0\linewidth]{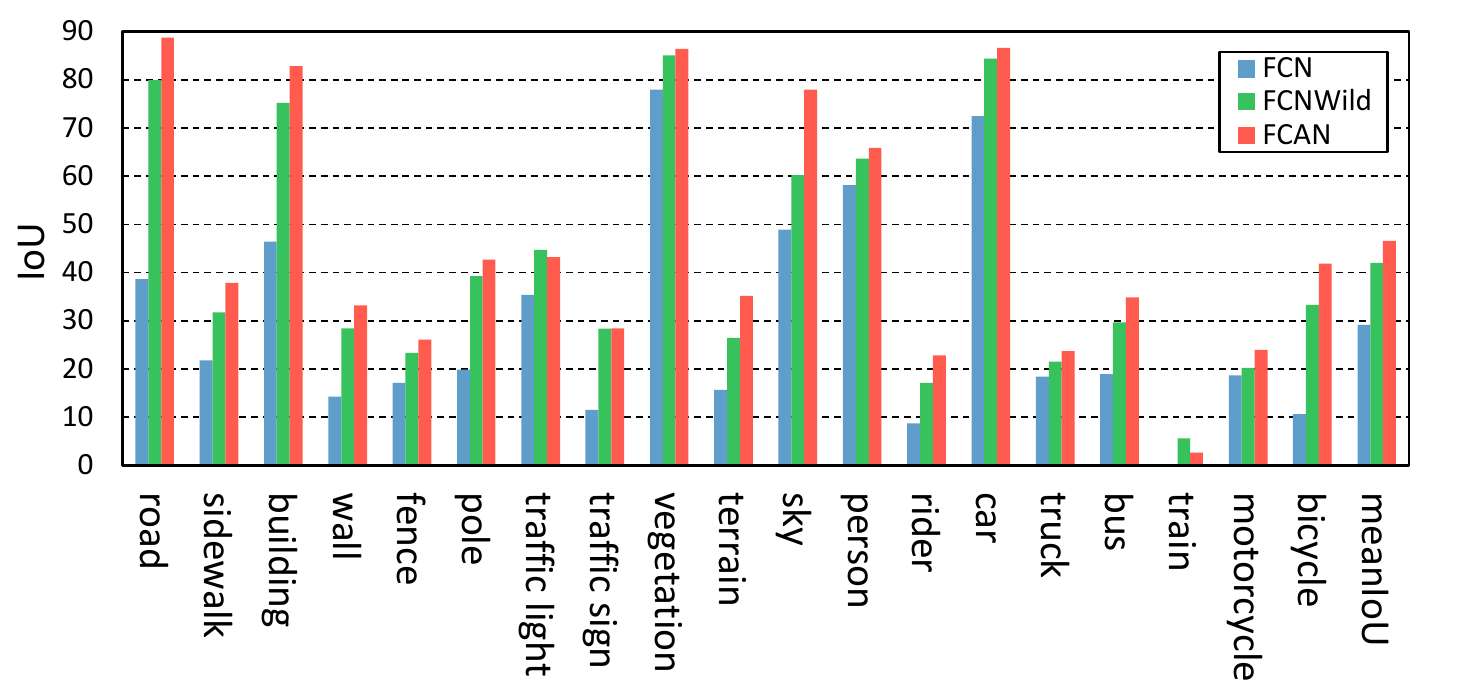}
    \end{center}
     \vspace{-0.12in}
    \caption{\small Per-category IoU performance of different approaches and mIoU performance averaged over all the 19 categories. }
    \label{fig:ResultCompareBars}
       \vspace{-0.08in}
\end{figure}

We compare with several state-of-the-art techniques. Domain Confusion \cite{tzeng2015simultaneous} (DC) aligns domains via domain confusion loss, which is optimized to learn a uniform distribution across different domains. Adversarial Discriminative Domain Adaptation \cite{tzeng2017adversarial} (ADDA) combines untied weight sharing and adversarial learning for discriminative feature learning.
FCNWild~\cite{BDDS_hoffman2016fcns} adopts fully convolutional adversarial training for domain adaptation on semantic segmentation. For fair comparison, the basic FCN utilized in all the methods are originated from ResNet-101. The performance comparisons are summarized in Table \ref{tab:state-of-the-art}. Compared to DC and ADDA in which domain discriminator are both devised on image level, FCNWild and FCAN performing domain-adversarial learning on region level exhibit better performance. Furthermore, FCAN by additionally incorporating ASPP strategy and reinforcing by AAN, leads to an apparent improvement over FCNWild. The multi-scale (MS) scheme boosts up the mIoU performance to 47.75\%. Figure \ref{fig:ResultCompareBars} details the performance across different categories. Our FCAN achieves the best performance in 17 out of 19 categories, which empirically validate the effectiveness of our model on category level.

\begin{figure}[t]
    \begin{center}
        \includegraphics[width=1.0\linewidth]{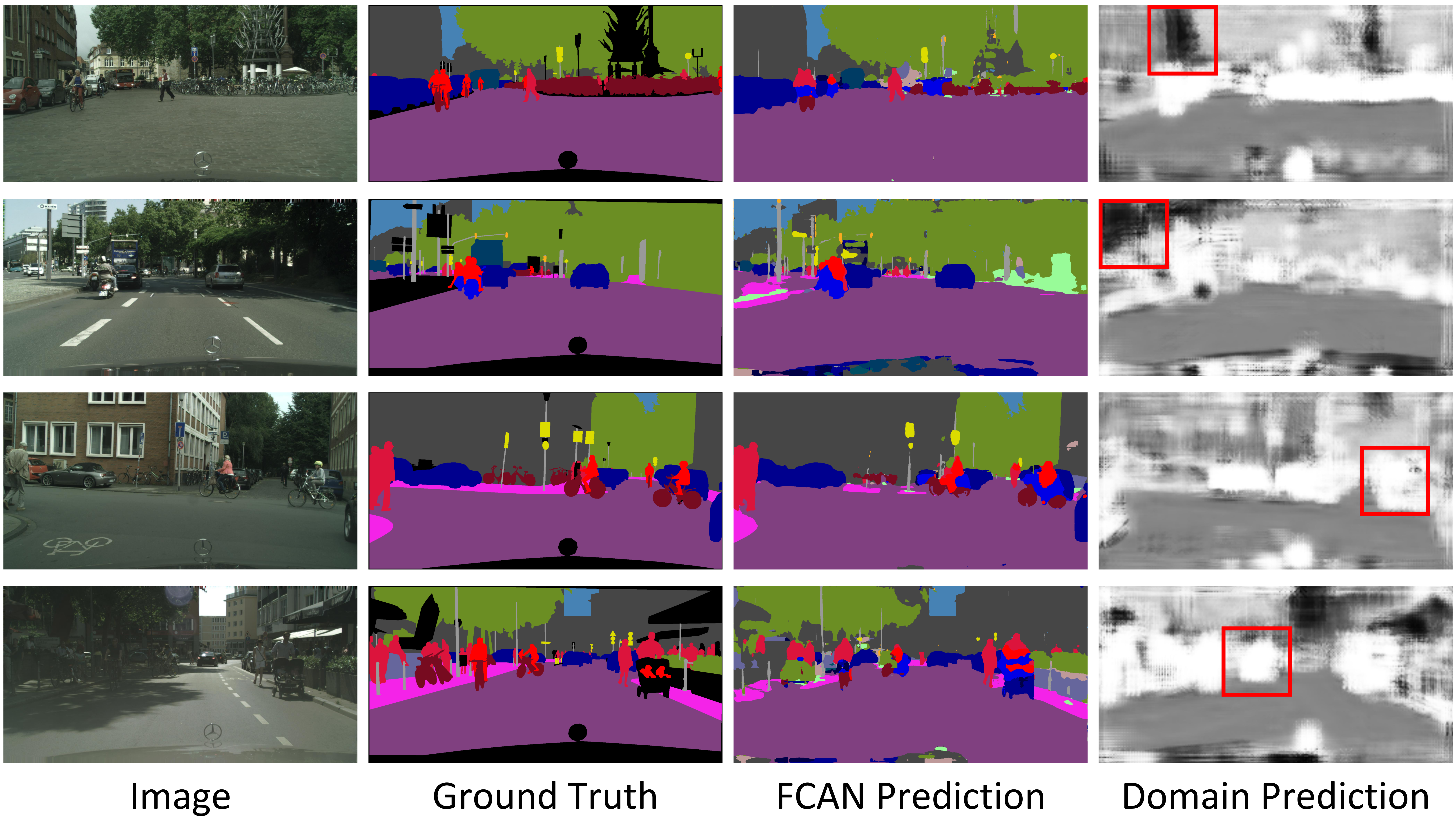}
    \end{center}
    \vspace{-0.1in}
    \caption{\small Examples of semantic segmentation results and the prediction maps by domain discriminator where brightness indicates the high probability of the region belonging to target domain.}
    \label{fig:Adapation Visual}
       \vspace{-0.15in}
\end{figure}

To examine domain discriminator learnt in FCAN, Figure \ref{fig:Adapation Visual} illustrates four image examples, including the original images, their ground truth, segmentation results by FCAN and prediction maps by domain discriminator. The brightness indicates that the region belongs to target domain with high probability. Let's recall that adversarial learning is to maximally fool the domain discriminator. That means ideally the prediction map of the images in target domain should be dark. For example, the domain discriminator predicts wrongly on the regions in the red bounding box in the first two images, which indicates that the representations on these regions tend to be indistinguishable. Hence, these regions (sky) are precisely segmented by FCAN. In contrast, domain discriminator predicts correctly on the regions in the last two images, indicating that the region representations are still domain-dependent. As such, the segmentation results on those regions (bicycle) are not that good.

\begin{table}
    \centering
    \small
    \caption{\small Results of Semi-supervised adaptation for Cityscapes.}
    \begin{tabular}{c|c|c} \hline
        \multirow{2}[0]{*}{\textbf{\# of images}} & \textbf{FCN}   & \textbf{FCAN} \\
        & (On Cityscapes) & (Semi-supervised) \\  \hline
        0     &   -   & 46.60 \\
        50    & 47.57 & 56.50 \\
        100   & 54.41 & 59.95   \\
        200   & 59.53 & 63.82   \\
        400   & 62.53 & 66.80  \\
        600   & 65.39 & 67.58  \\
        800   & 67.01 & 68.42 \\
        1000  & 68.05 & 69.17 \\ \hline
        \end{tabular}
        \label{tab:SS}
        \vspace{-0.07in}
\end{table}

\subsection{Semi-Supervised Adaptation}
Another common scenario in practice is that there is a small number of labeled training examples in target domain. Hence, we extend our FCAN to a semi-supervised version, which takes the training set of Cityscapes as labeled data {\small $\mathcal{X}_{t}^{l}$}. Technically, the pixel-level classification loss on images in target domain is further taken into account and the overall objective in Eq.(\ref{loss_adv_3}) then changes to {\small $\max_{F}\min_{D}    \left\{    \mathcal{L}_{adv}(\mathcal{X}_{s},\mathcal{X}_{t}) - \lambda_{s}\mathcal{L}_{seg}(\mathcal{X}_{s}) - \lambda_{t}\mathcal{L}_{seg}(\mathcal{X}_{t}^{l}) \right\}$}. Table \ref{tab:SS} shows the mIoU performances with the increase of labeled training data from target domain. It is also worth noting that here FCN is directly learnt on the labeled data in target domain and FCAN refers to our semi-supervised version. As expected, the performance gain of FCAN tends to be large if only a few hundred images in target domain are included in training. The gain is gradually decreased when increasing the number of images from Cityscapes. Even when the number reaches 1$k$, our semi-supervised FCAN is still slightly better than supervised FCN.

\begin{table}
    \centering
    \small
    \caption{\small Comparisons of different unsupervised domain adaptation methods on BDDS.}
    \begin{tabular}{l|c} \hline
        \textbf{Method} & \textbf{mIoU} \\ \hline
        FCNWild~\cite{BDDS_hoffman2016fcns}          & 39.37 \\ \hline \hline
        FCAN                                         & 43.35 \\
        FCAN(MS)                                     & 45.47 \\
        FCAN(MS+EN)                                  & 47.53 \\ \hline
    \end{tabular}
    \label{tab:BDDS}
    \vspace{-0.18in}
\end{table}

\subsection{Results on BDDS}
\begin{figure}[t]
    \setlength{\abovecaptionskip}{0pt}
    \begin{center}
        \includegraphics[width=1.0\linewidth]{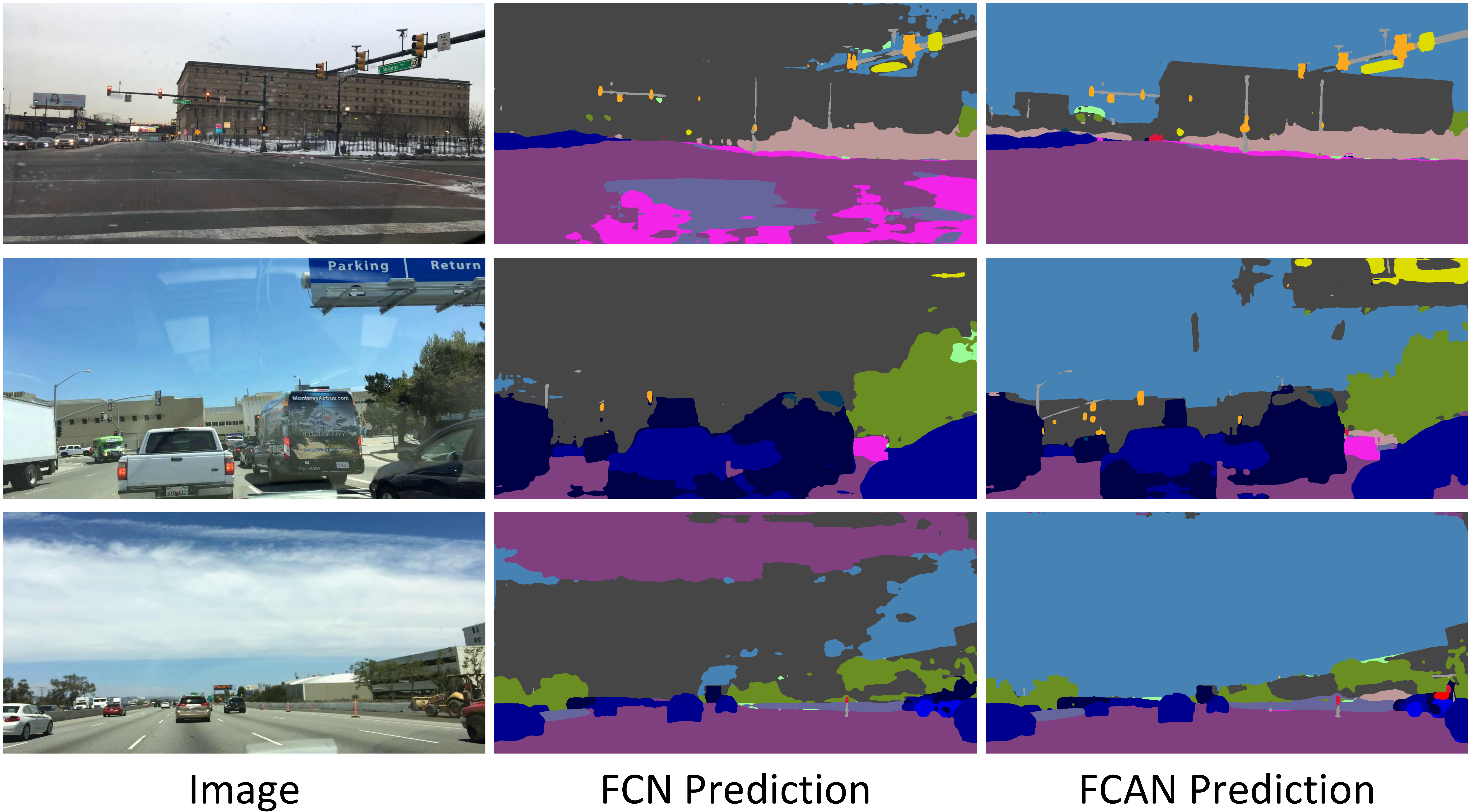}
    \end{center}
    \vspace{-0.05in}
    \caption{\small Examples of semantic segmentation results in BDDS.}
    \label{fig:BDDS_compare}
    \vspace{-0.15in}
\end{figure}

In addition to Cityscapes dataset, we also take BDDS as target domain to evaluate the unsupervised setting of our FCAN. The performance comparisons are summarized in Table \ref{tab:BDDS}. In particular, the mIoU performance of FCAN achieves 43.35\%, making the improvement over FCNWild by 3.98\%. The multi-scale setting, i.e., FCAN(MS), increases the performance to 45.47\%. Finally, the ensemble version FCAN(MS+EN) by fusing the models derived from ResNet-101, ResNet-152 and SENet \cite{senet_hu2017}, could boost up the mIoU to 47.53\%. Figure \ref{fig:BDDS_compare} shows three semantic segmentation examples in BDDS, which are output by FCN and FCAN, respectively. Clearly, FCAN obtains much more promising segmentation results. Even in the case of a reflection (second row) or patches of cloud (third row) in the sky, our FCAN can segment the sky well.

\section{Conclusion}
We have presented Fully Convolutional Adaptation Networks (FCAN) architecture, which explores domain adaptation for semantic segmentation. Particularly, we study the problem from the viewpoint of both visual appearance-level and representation-level adaptation. To verify our claim, we have devised Appearance Adaptation Networks (AAN) and Representation Adaptation Networks (RAN) respectively in our FCAN for each purpose. AAN is to render an image in one domain with the domain ``style" from the other one, resulting in invariant appearance across two domains. RAN aims to guide the representation learning in a domain-adversarial manner, which ideally outputs domain-invariant representations. Experiments conducted on the transfer from game videos (GTA5) to urban street-view scenes (Cityscapes) validate our proposal and analysis. More remarkably, we achieve new state-of-the-art performances when transferring game videos to drive-cam videos (BDDS). Our possible future works include two directions. First, more advanced techniques of rendering the semantic content of an image with another statistical pattern will be investigated in AAN. Second, we will further extend our FCAN to other specific segmentation scenarios, e.g., indoor scenes segmentation or portrait segmentation, where the synthetic data could be easily produced.

{\small
\bibliographystyle{ieee}
\bibliography{egbib}
}

\end{document}